\documentclass[10pt,twocolumn,letterpaper]{article}

\usepackage{cvpr}
\usepackage{times}
\usepackage{epsfig}
\usepackage{graphicx}
\usepackage{amsmath}
\usepackage{amssymb}

\usepackage{multirow}

\usepackage{booktabs}
\usepackage{balance}

\usepackage[pagebackref=true,breaklinks=true,letterpaper=true,colorlinks,bookmarks=false]{hyperref}
\cvprfinalcopy 

\def\cvprPaperID{685} 

\ifcvprfinal\fi
\begin{document}


\title{Rethinking Classification and Localization for Object Detection}

\author{Yue Wu\textsuperscript{1}, Yinpeng Chen\textsuperscript{2}, Lu Yuan\textsuperscript{2}, Zicheng Liu\textsuperscript{2}, Lijuan Wang\textsuperscript{2}, Hongzhi Li\textsuperscript{2} and Yun Fu\textsuperscript{1} \\
	\textsuperscript{1}Northeastern University, \textsuperscript{2}Microsoft\\
{\tt\small \{yuewu,yunfu\}@ece.neu.edu, \{yiche,luyuan,zliu,lijuanw,hongzhi.li\}@microsoft.com} 
}

\maketitle

\begin{abstract}
Two head structures (i.e. fully connected head and convolution head) have been widely used in R-CNN based detectors for classification and localization tasks. However, there is a lack of \textbf{understanding} of how does these two head structures work for these two tasks. To address this issue, we perform a thorough analysis and find an interesting fact that the two head structures have \textbf{opposite} preferences towards the two tasks. Specifically, the fully connected head (\textit{fc-head}) is more suitable for the classification task, while the convolution head (\textit{conv-head}) is more suitable for the localization task. 
Furthermore, we examine the output feature maps of both heads and find that \textit{fc-head} has more spatial sensitivity than \textit{conv-head}.
Thus, \textit{fc-head} has more capability to distinguish a complete object from part of an object, but is not robust to regress the whole object.
Based upon these findings, we propose a \textbf{Double-Head} method, which has a fully connected head focusing on classification and a convolution head for bounding box regression. Without bells and whistles, our method gains +3.5 and +2.8 AP on MS COCO dataset from Feature Pyramid Network (FPN) baselines with ResNet-50 and ResNet-101 backbones, respectively.
\end{abstract}

\section{Introduction}

\begin{figure}[t]
	\begin{center}
		\includegraphics[width=0.9\linewidth]{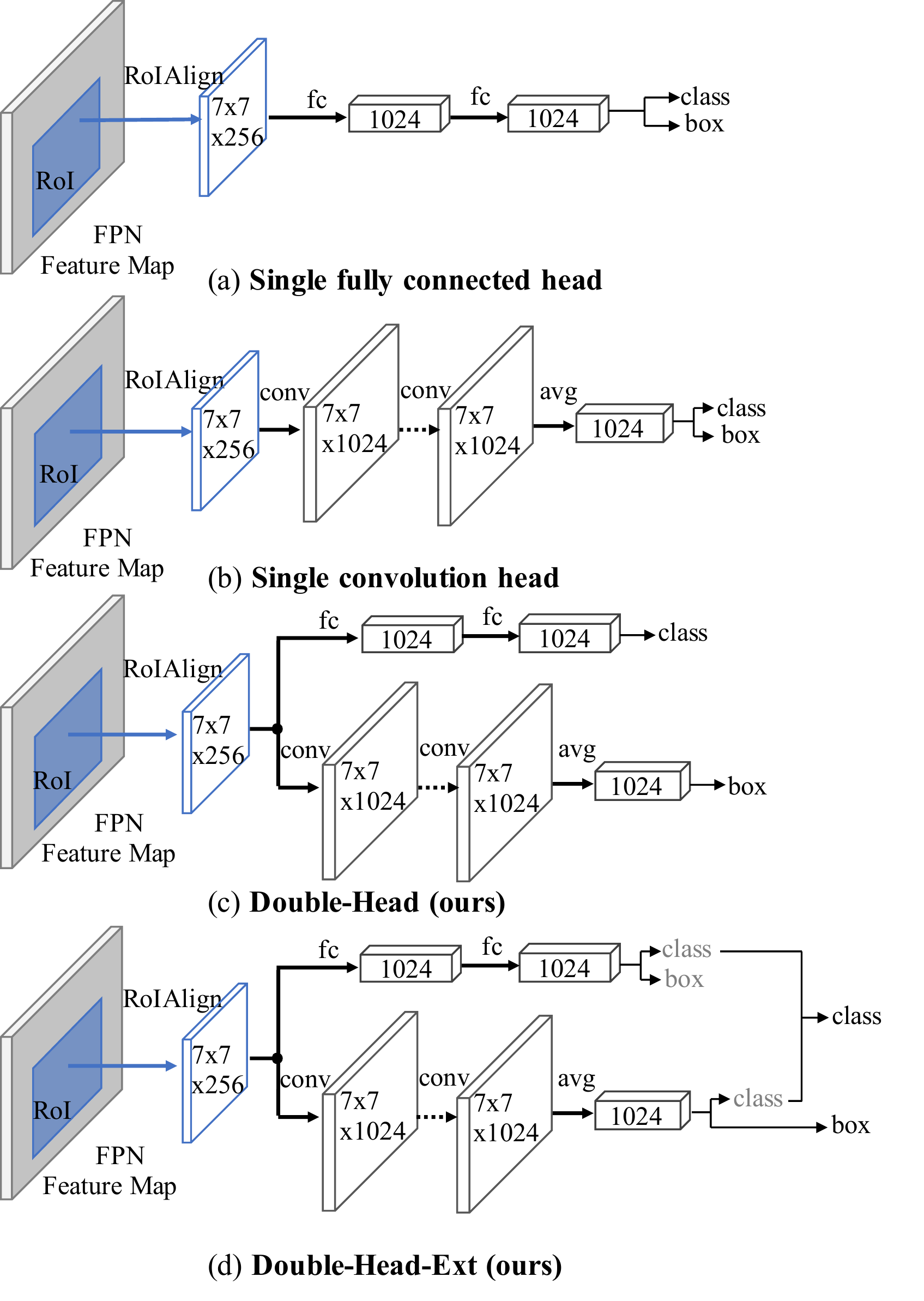}
	\end{center}
	\vspace{-2mm}
	\caption{Comparison between single head and double heads, (a) a single fully connected (2-\textit{fc}) head, (b) a single convolution head, (c) Double-Head, which splits classification and localization on a fully connected head and a convolution head respectively, and (d) Double-Head-Ext, which extends Double-Head by introducing supervision from unfocused tasks during training and combining classification scores from both heads during inference.}
	\label{fig:overview}
	\vspace{-4mm}
\end{figure}

Most two-stage object detectors \cite{girshick15fastrcnn, girshick2014rcnn, ren2015faster, Dai_RFCN, Lin_FPN} share a head for both classification and bounding box regression.
Two different head structures are widely used. Faster R-CNN \cite{ren2015faster} uses a convolution head (conv5) on a single level feature map (conv4), while FPN \cite{Lin_FPN} uses a fully connected head (2-\textit{fc}) on multiple level feature maps. However, there is a lack of \textbf{understanding} between the two head structures with respect to the two tasks (object classification and localization).

In this paper, we perform a thorough comparison between the fully connected head (\textit{fc-head}) and the convolution head (\textit{conv-head}) on the two detection tasks, i.e. object classification and localization. We find that \textit{these two different head structures are complementary}. \textit{fc-head} is more suitable for the classification task as its classification score is more correlated to the intersection over union (IoU) between a proposal and its corresponding ground truth box. Meanwhile, \textit{conv-head} provides more accurate bounding box regression.

We believe this is because \textit{fc-head} is spatially sensitive, having different parameters for different parts of a proposal, while \textit{conv-head} shares convolution kernels for all parts. 
To validate this, we examine the output feature maps of both heads and confirm that \textit{fc-head} is more spatially sensitive.
As a result, \textit{fc-head} is better to distinguish between a complete object and part of an object (classification) and \textit{conv-head} is more robust to regress the whole object (bounding box regression). 


In light of above findings, we propose a {Double-Head} method, which includes a fully connected head (\textit{fc-head}) for classification and a convolution head  (\textit{conv-head}) for bounding box regression (see Figure \ref{fig:overview}-(c)), to leverage advantages of both heads.
This design outperforms both single \textit{fc-head} and single \textit{conv-head} (see Figure \ref{fig:overview}-(a), (b)) by a  non-negligible margin. 
In addition, we extend Double-Head (Figure \ref{fig:overview}-(d)) to further improve the accuracy by leveraging unfocused tasks (i.e. classification in \textit{conv-head}, and bounding box regression in \textit{fc-head}).  
Our method outperforms FPN baseline by a non-negligible margin on MS COCO 2017 dataset \cite{lin2014microsoft}, gaining 3.5 and 2.8 AP for using ResNet-50 and ResNet-101 backbones, respectively. 

\section{Related Work}
\noindent \textbf{One-stage Object Detectors:} 
OverFeat \cite{sermanet2013overfeat} detects objects by sliding windows on feature maps. 
SSD \cite{liu2016ssd,fu2017dssd} and YOLO \cite{redmon2016you,redmon2017yolo9000,redmon2018yolov3} have been tuned for speed by predicting object classes and locations directly. RetinaNet \cite{lin2018focal} alleviates the extreme foreground-background class imbalance problem by introducing focal loss. Point-based methods \cite{Law2018cornernet, Law2019cornernetlite, Zhou2019objaspt, Duan2019centernet, Zhou2019extremenet} model an object as keypoints (corner, center, etc), and are built on  keypoint estimation networks.

\noindent  \textbf{Two-stage Object Detectors:} RCNN \cite{girshick2014rich} applies a deep neural network to extract features from proposals generated by selective search \cite{uijlings2013selective}. 
SPPNet \cite{he2014spatial} speeds up RCNN significantly using spatial pyramid pooling. 
Fast RCNN \cite{girshick15fastrcnn} 
improves the speed and performance utilizing a differentiable RoI Pooling.
Faster RCNN  \cite{ren2015faster} introduces Region Proposal Network (RPN) to generate proposals. R-FCN \cite{Dai_RFCN} employs position sensitive RoI pooling to address the  translation-variance problem.
FPN \cite{Lin_FPN} builds a top-down architecture with lateral connections to extract features across multiple layers. 

\noindent  \textbf{Backbone Networks:} 
Fast RCNN \cite{girshick15fastrcnn} and Faster RCNN \cite{ren2015faster} extract features from conv4 of VGG-16 \cite{simonyan2014very}, while
FPN \cite{Lin_FPN} utilizes features from multiple layers (conv2 to conv5) of ResNet \cite{he2016deep}. 
Deformable ConvNets \cite{dai2017deformable,zhu2018deformable} propose deformable convolution and deformable Region of Interests (RoI) pooling to augment spatial sampling locations. 
Trident Network \cite{li2019scale} generates scale-aware feature maps with multi-branch architecture. MobileNet \cite{howard2017mobilenets,sandler2018mobilenetv2} and ShuffleNet \cite{zhang2018shufflenet,ma2018shufflenet} introduce efficient operators (like depth-wise convolution, group convolution, channel shuffle, etc) to speed up on mobile devices.

\noindent  \textbf{Detection Heads:} 
Light-Head RCNN \cite{li2017light} introduces an efficient head network with thin feature maps.
Cascade RCNN \cite{Cai_2018_CVPR} constructs a sequence of detection heads trained with increasing IoU thresholds. Feature Sharing Cascade RCNN \cite{li2019rethinking} utilizes feature sharing to ensemble multi-stage outputs from Cascade RCNN \cite{Cai_2018_CVPR} to improve the results.
Mask RCNN \cite{he2017mask} introduces an extra head for instance segmentation.  
COCO Detection 18 Challenge winner (Megvii) \cite{megviiCOCO2018} couples bounding box regression and instance segmentation in a convolution head.
IoU-Net \cite{jiang2018acquisition} introduces a branch to predict IoUs between detected bounding boxes and their corresponding ground truth boxes.
Similar to IoU-Net, Mask Scoring RCNN \cite{huang2019msrcnn} presents an extra head to predict Mask IoU scores for each segmentation mask. 
He et. al. \cite{he2019bounding} learns uncertainties of bounding box prediction with an extra task to improve the localization results.  
Learning-to-Rank \cite{Tan_2019_ICCV} utilizes an extra head to produce a rank value of a proposal for Non-Maximum Suppression (NMS). 
Zhang and Wang \cite{Zhang_2019_ICCV} point out that there exist mis-alignments between classification and localization task domains. 
In contrast to existing methods, which apply a single head to extract Region of Interests (RoI) features for both classification and bounding box regression tasks, we propose to split these two tasks into different heads, based upon our thorough analysis.

\section{Analysis: Comparison between \textit{fc-head} and \textit{conv-head}}  \label{sec:analysisTwoHeads}
In this section, we compare \textit{fc-head} and \textit{conv-head} for both classification and bounding box regression. For each head, we train a model with FPN backbone \cite{Lin_FPN} using ResNet-50 \cite{he2016deep} on MS COCO 2017 dataset \cite{lin2014microsoft}. The \textit{fc-head} includes two fully connected layers. The \textit{conv-head} has five residual blocks. The evaluation and analysis is conduct on the MS COCO 2017 validation set with 5,000 images. 
\textit{fc-head} and \textit{conv-head} have 36.8\% and 35.9\% AP, respectively. 

\subsection{Data Processing for Analysis} \label{sec:slidingwindowclassification}
To make a fair comparison, we perform analysis for both heads on predefined proposals rather than proposals generated by RPN \cite{ren2015faster}, as the two detectors have different proposals. The predefined proposals include sliding windows around the ground truth box with different sizes. 
For each ground truth object, we generate about 14,000 proposals. The IoUs between these proposals and the ground truth box (\textit{denoted as proposal IoUs}) gradually change from zero (background) to one (the ground truth box). For each proposal, both detectors (\textit{fc-head} and \textit{conv-head}) generate classification scores and regressed bounding boxes.This process is applied for all objects in the validation set.

We split the IoUs between predefined proposals and their corresponding ground truth into 20 bins uniformly, and group these proposals accordingly. For each group, we calculate mean and standard deviation of classification scores and IoUs of regressed boxes. 
Figure \ref{fig:slidingwindowMeanstd} shows the results for small, medium and large objects.

\subsection{Comparison on Classification Task}  \label{sec:fcConvCls}
The first row of Figure \ref{fig:slidingwindowMeanstd} shows the classification scores for both \textit{fc-head} and \textit{conv-head}.
Compared to \textit{conv-head}, \textit{fc-head} provides higher scores for proposals with higher IoUs. 
This indicates that \textit{classification scores of \textit{fc-head} are more correlated to  IoUs between proposals and corresponding ground truth than of \textit{conv-head}}, especially for small objects.
To validate this, we compute the Pearson correlation coefficient (PCC) between proposal IoUs and classification scores. The results (shown in Figure \ref{fig:slidingwindowPscores} (Left)) demonstrate that the classification scores of \textit{fc-head} are more correlated to the proposal IoUs.

We also compute Pearson correlation coefficient for the proposals generated by RPN \cite{ren2015faster} and final detected boxes after NMS. Results are shown in 
Figure \ref{fig:slidingwindowPscores} (Right). 
Similar to the predifined proposals, \textit{fc-head} has higher PCC than \textit{conv-head}. 
Thus, the detected boxes with higher IoUs are ranked higher when calculating AP due to their higher classification scores.

\subsection{Comparison on Localization Task}
\begin{figure}[t]
	\begin{center}
        \includegraphics[width=\linewidth]{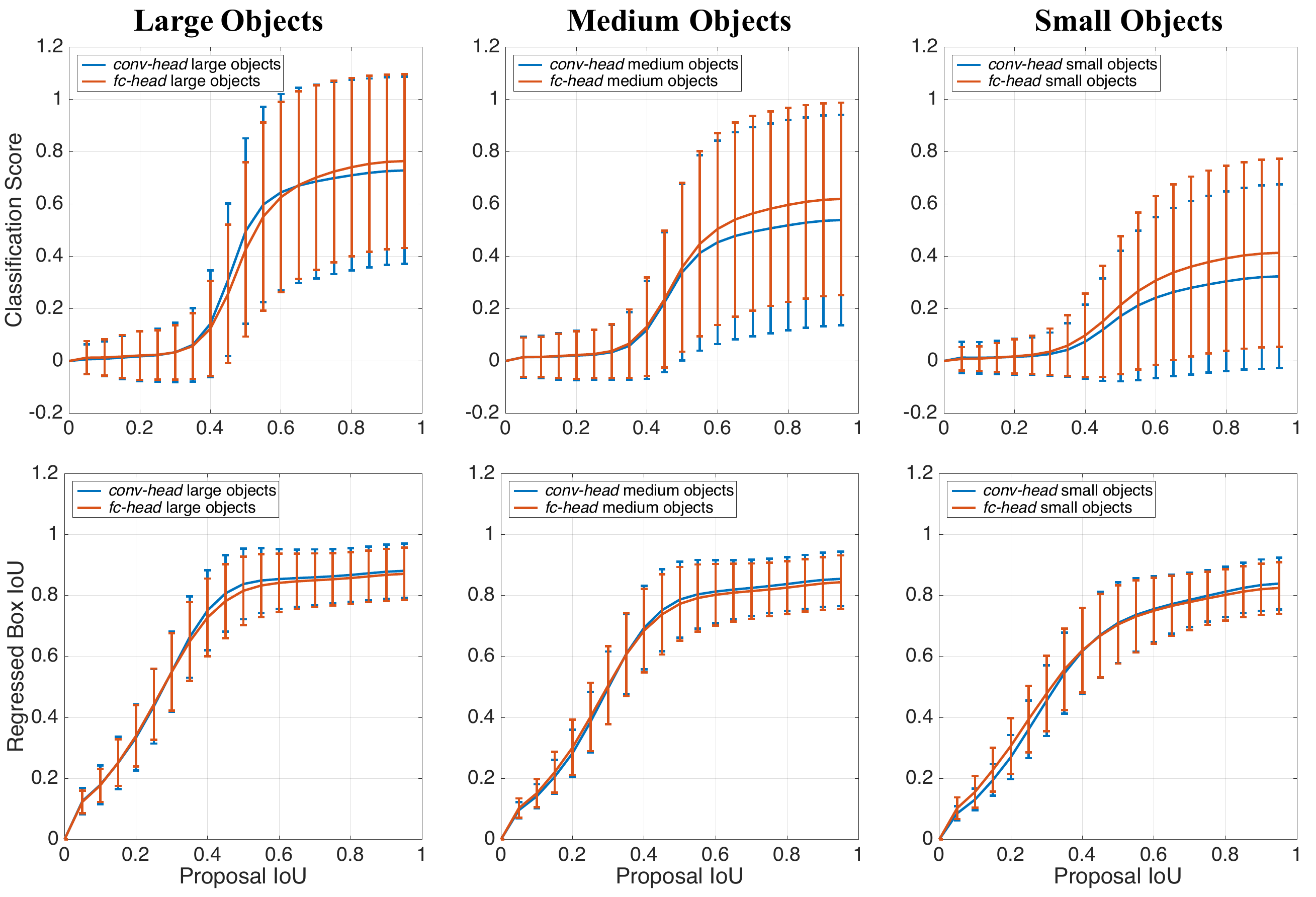}
	\end{center}
	\vspace{-2mm}
	\caption{
	Comparison between \textit{fc-head} and \textit{conv-head}.
	Top row: mean and standard deviation of classification scores. Bottom row: mean and standard deviation of IoUs between regressed boxes and their corresponding ground truth.
	Classification scores in \textit{fc-head} are more correlated to proposal IoUs than in \textit{conv-head}. \textit{conv-head} has better regression results than \textit{fc-head}. 
}
	\vspace{-2mm}
	\label{fig:slidingwindowMeanstd}
\end{figure}

\begin{figure}[t]
	\begin{center}
		\includegraphics[width=0.9\linewidth]{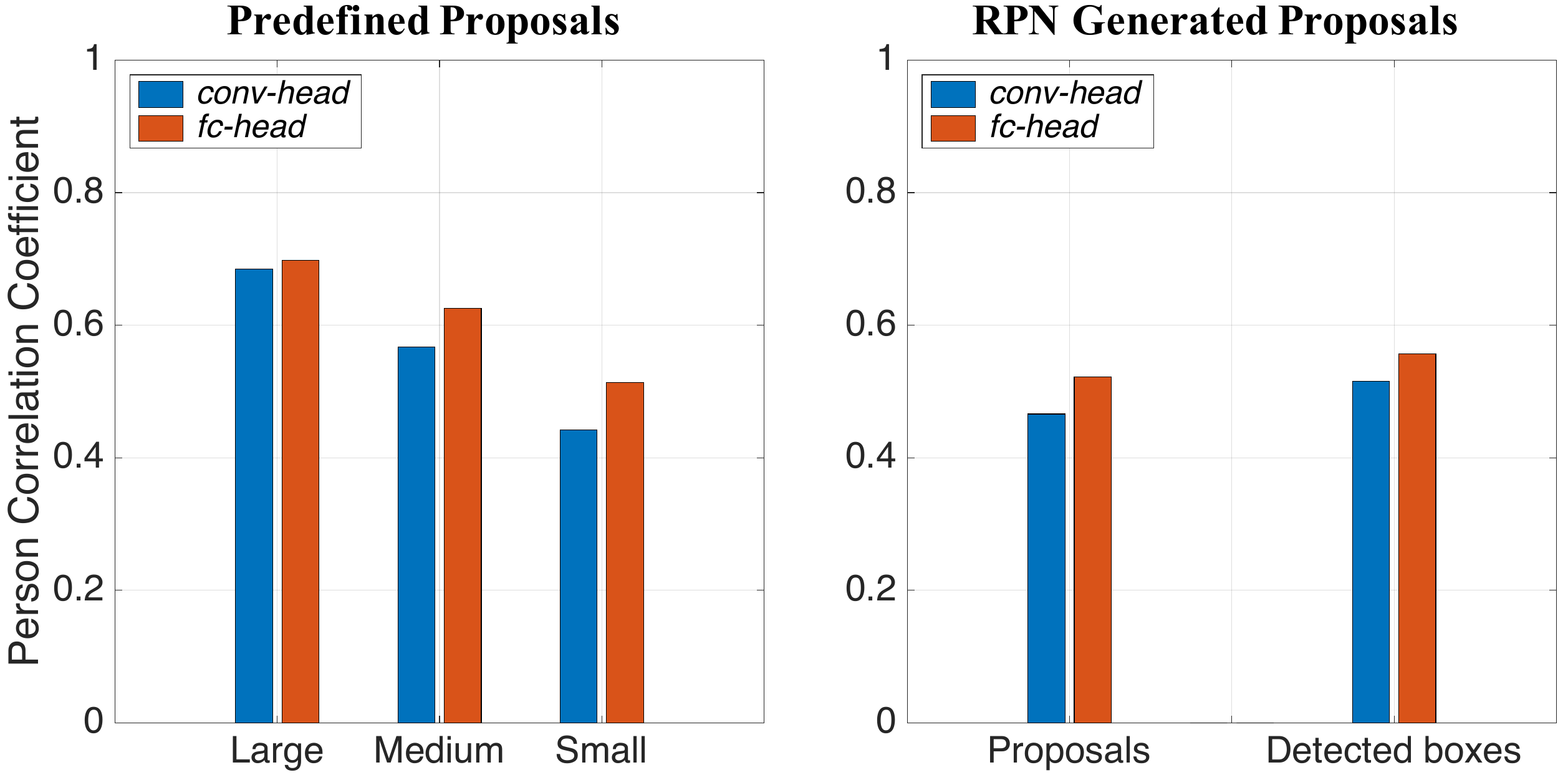}
	\end{center}
	\vspace{-2mm}
	\caption{
Pearson correlation coefficient (PCC) between classification scores and IoUs.
Left: PCC of predefined proposals for large, medium and small objects. 
Right: PCC of proposals generated by RPN and detected boxes after NMS.
	}
	\label{fig:slidingwindowPscores}
\end{figure}

The second row of Figure \ref{fig:slidingwindowMeanstd} shows IoUs between the regressed boxes and their corresponding ground truth for both \textit{fc-head} and \textit{conv-head}.
Compared to \textit{fc-head}, the regressed boxes of \textit{conv-head} are more accurate when the proposal IoU is above 0.4.
This demonstrates that \textit{\textit{conv-head} has better regression ability than \textit{fc-head}}.




\begin{figure}[t]
	\begin{center}
		\includegraphics[width=\linewidth]{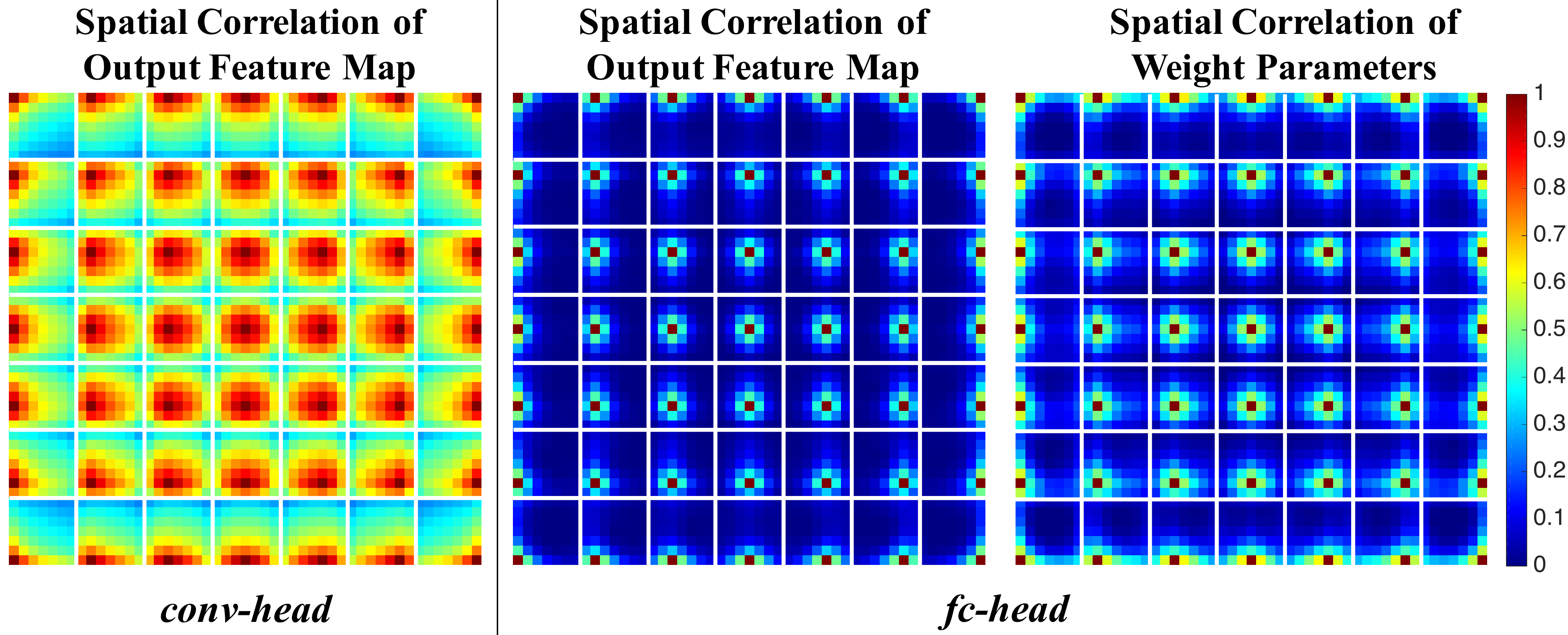}
	\end{center}
	\vspace{-2mm}
	\caption{
	Left: 	Spatial correlation in output feature map of \textit{conv-head}. Middle: 	Spatial correlation in output feature map of \textit{fc-head}. Right: Spatial correlation in weight parameters of \textit{fc-head}.
	\textit{conv-head} has significantly more spatial correlation in output feature map  than \textit{fc-head}. 
	\textit{fc-head} has a similar spatial correlation pattern in output feature map and weight parameters. 
	}
	\label{fig:fcspatial}
\end{figure}
\subsection{Discussion}
\textit{Why does \textit{fc-head} show more correlation between the classification scores and proposal IoUs, and perform worse in localization?} 
We believe it is because \textit{fc-head} is more spatially sensitive than \textit{conv-head}. 
Intuitively, \textit{fc-head} applies \textit{unshared} transformations (fully connected layer) over different positions of the input feature map. Thus, the spatial information is implicitly embedded. 
The spatial sensitivity of \textit{fc-head} helps distinguish between a complete object and part of an object, but is not robust to determine the offset of the whole object.
In contrast, \textit{conv-head} uses a \textit{shared} transformation (convolutional kernels) on all positions of the input feature map, and uses average pooling to aggregate.


Next, we inspect the spatial sensitivity of \textit{conv-head} and \textit{fc-head}.
For \textit{conv-head}  whose output feature map is a $7\times7$ grid, we compute the spatial correlation between any pair of locations using the cosine distance between the corresponding two feature vectors. 
This results in a $7\times7$ correlation matrix per cell, representing the correlation between the current cell and other cells. 
Thus, the spatial correlation of a output feature map can be visualized by tiling the correlation matrices of all cells in a $7\times7$ grid.
Figure \ref{fig:fcspatial} (Left) shows the average spatial correlation of \textit{conv-head} over multiple objects.
For \textit{fc-head} whose output is not a feature map, but a feature vector with dimension 1024, we reconstruct its output feature map. 
This can be done by splitting the weight matrix of fully connected layer ($256\cdot 7\cdot 7 \times 1024$) by spatial locations. Each cell in the $7\times7$ grid has a transformation matrix with dimension $256\times1024$, which is used to generate output features for that cell.
Thus, output feature map $7\times7\times1024$ for \textit{fc-head} is reconstructed. Then we can compute its spatial correlation in a similar manner to \textit{conv-head}.
Figure \ref{fig:fcspatial} (Middle) shows the average spatial correlation in output feature map of \textit{fc-head} over multiple objects.
\textit{fc-head} has significant less spatial correlation than \textit{conv-head}. This supports our conjecture that \textit{fc-head} is more spatially sensitive than \textit{conv-head}, making it easier to distinguish if one proposal covers one complete or partial object. On the other hand, it is not as robust as \textit{conv-head} to regress bounding boxes.

We further examine the spatial correlation of weight parameters ($256\cdot 7\cdot 7 \times 1024$) in \textit{fc-head}, by splitting them along spatial locations. As a result, each cell of the $7\times7$ grid has a matrix with dimension $256\times1024$, which is used to compute correlation with other cells.
Similar to the correlation analysis on output feature map, we compute the correlation matrices for all cells.
Figure \ref{fig:fcspatial} (Right) shows the spatial correlation in weight parameters of \textit{fc-head}. It has a similar pattern to the spatial correlation in output feature map of \textit{fc-head} (shown in Figure \ref{fig:fcspatial} (Middle)).
\section{Our Approach: Double-Head}
Based upon above analysis, we propose a double-head method to leverage the advantages of two head structures.
In this section, 
we firstly introduce the network structure of Double-Head,
 which has a fully connected head (\textit{fc-head}) for classification and a convolution head (\textit{conv-head}) for bounding box regression. 
Then, 
we extend Double-Head to Double-Head-Ext 
by leveraging unfocused tasks (i.e. bounding box regression in \textit{fc-head} and classification in \textit{conv-head}).

\subsection{Network Structure}\label{sec:our-net}
Our Double-Head method (see Figure \ref{fig:overview}-(c)) splits classification and localization into \textit{fc-head} and \textit{conv-head}, respectively. The details of backbone and head networks are described as follows: 

\noindent \textbf{Backbone:}
We use FPN \cite{Lin_FPN} backbone to generate region proposals and extract object features from multiple levels using RoIAlign \cite{he2017mask}. Each proposal has a feature map with size $256\times7\times7$, which is transformed  by \textit{fc-head} and \textit{conv-head} into two feature vectors (each with dimension 1024) for classification and bounding box regression, respectively. 

\noindent\textbf{Fully Connected Head (\textit{fc-head})}
has two fully connected layers (see Figure \ref{fig:overview}-(c)), following the design in FPN \cite{Lin_FPN} (Figure \ref{fig:overview}-(a)). The output dimension is 1024. The parameter size is 13.25M.

\begin{figure}[t]
	\begin{center}
		\includegraphics[width=\linewidth]{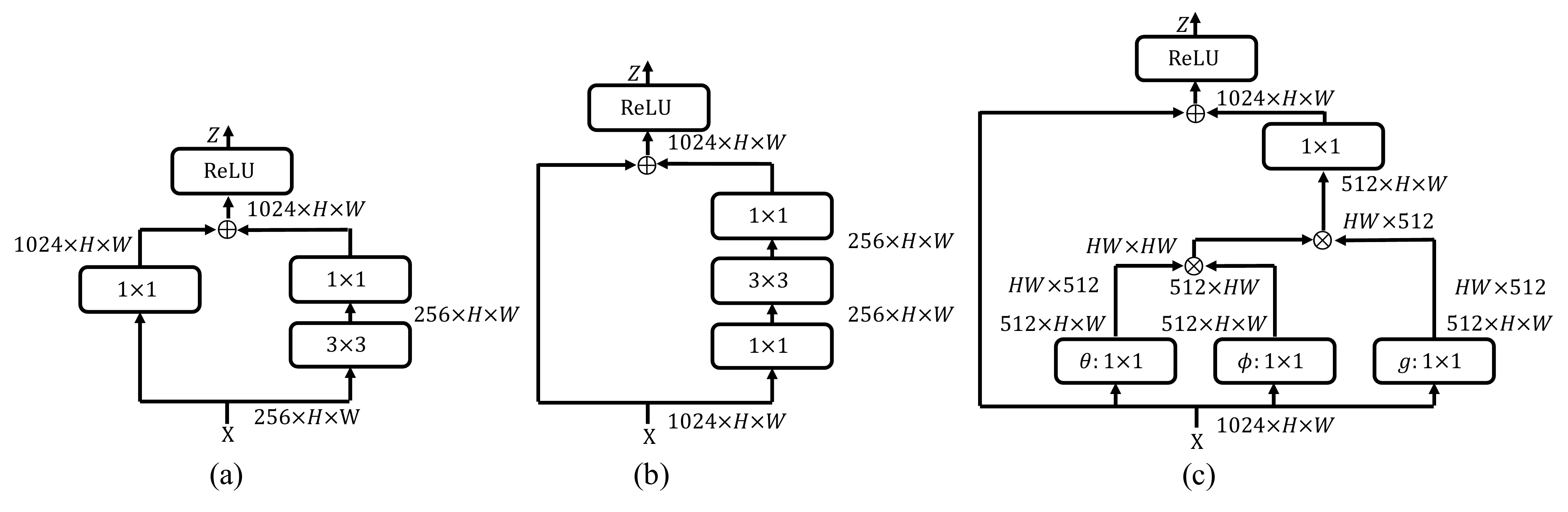}
	\end{center}
	\vspace{-2mm}
	\caption{Network architectures of three components: (a) residual block to increase the number of channels (from 256 to 1024), (b) residual bottleneck block, and (c) non-local block.}
	\label{fig:upchannels}
\end{figure}

\label{sec:convheadstrcture} 
\noindent\textbf{Convolution Head (\textit{conv-head})} stacks $K$ residual blocks \cite{he2016deep}. The first block increases the number of channels from 256 to 1024 (shown in Figure \ref{fig:upchannels}-(a)), and others are bottleneck blocks \cite{he2016deep} (shown in Figure \ref{fig:upchannels}-(b)). At the end, average pooling is used to generate the feature vector with dimension 1024. Each residual block has 1.06M parameters.
We also introduce a variation for the convolution head by inserting a non-local block \cite{wang2018non} (see Figure \ref{fig:upchannels}-(c)) before each bottleneck block to enhance foreground objects. Each non-local block has 2M parameters.

\noindent\textbf{Loss Function:} Both heads (i.e. \textit{fc-head} and \textit{conv-head}) are jointly trained with region proposal network (RPN) end to end. The overall loss is computed as follows:
\begin{align}
\mathcal{L} &= \omega^{fc}\mathcal{L}^{fc} + \omega^{conv}\mathcal{L}^{conv} + \mathcal{L}^{rpn},
\label{eq:loss-det}
\end{align}
where $\omega^{fc}$ and $\omega^{conv}$ are weights for \textit{fc-head} and \textit{conv-head}, respectively. $\mathcal{L}^{fc}$, $\mathcal{L}^{conv}$, $\mathcal{L}^{rpn}$ are the losses for \textit{fc-head}, \textit{conv-head} and RPN, respectively.

\subsection{Extension: Leveraging Unfocused Tasks}
In vanilla Double-Head, each head focuses on its assigned task (i.e. classification in \textit{fc-head} and bounding box regression in \textit{conv-head}). In addition, we found that unfocused tasks (i.e. bounding box regression in \textit{fc-head} and classification in \textit{conv-head}) are helpful in two aspects: (a) bounding box regression provides auxiliary supervision for \textit{fc-head}, and (b) classifiers from both heads are complementary. Therefore, we introduce unfocused task supervision in training and propose a complementary fusion method to combine classification scores from both heads during inference (see Figure \ref{fig:overview}-(d)). This extension is referred to as Double-Head-Ext.

\noindent \textbf{Unfocused Task Supervision}: Due to the introduction of the unfocused tasks, the loss for \textit{fc-head} ($\mathcal{L}^{fc}$) includes both classification loss and bounding box regression loss as follows:
\begin{align}
\mathcal{L}^{fc} &= \lambda^{fc}L^{fc}_{cls}+(1-\lambda^{fc})L^{fc}_{reg},
\label{eq:loss-heads}
\end{align}
where $L^{fc}_{cls}$ and $L^{fc}_{reg}$ are the classification and bounding box regression losses in \textit{fc-head}, respectively.  $\lambda^{fc}$ is the weight that controls the balance between the two losses in \textit{fc-head}.
%
In the similar manner, we define the loss for the convolution head ($\mathcal{L}^{conv}$) as follows:
\begin{align}
\mathcal{L}^{conv} &= (1-\lambda^{conv})L^{conv}_{cls}+\lambda^{conv}L^{conv}_{reg},
\label{eq:loss-heads-2}
\end{align}
where $L^{conv}_{cls}$ and $L^{conv}_{reg}$ are classification and bounding box regression losses in \textit{conv-head}, respectively. 
Different from $\lambda^{fc}$ that is multiplied by the classification loss $L^{fc}_{cls}$, 
the balance weight $\lambda^{conv}$ is multiplied by the regression loss $L^{conv}_{reg}$, as the bounding box regression is the focused task in \textit{conv-head}. Note that the vanilla Double-Head is a special case when $\lambda^{fc}=1$ and $\lambda^{conv}=1$.
Similar to FPN \cite{Lin_FPN},
cross entropy loss is applied to classification, and Smooth-$L_1$ loss is used for bounding box regression.

\noindent \textbf{Complementary Fusion of Classifiers}: 
We believe that the two heads (i.e. \textit{fc-head} and \textit{conv-head}) capture complementary information for object classification due to their different structures. Therefore we propose to fuse the two classifiers as follows:
\begin{align}
s &= s^{fc} + s^{conv}(1-s^{fc}) = s^{conv} + s^{fc}(1-s^{conv}),
\label{eq:cls-fusion}
\end{align}
where $s^{fc}$ and $s^{conv}$ are classification scores from \textit{fc-head} and \textit{conv-head}, respectively. The increment from the first score (e.g. $s^{fc}$) is a product of the second score and the reverse of the first score (e.g. $s^{conv}(1-s^{fc})$). This is different from \cite{Cai_2018_CVPR} which combining all classifiers by average. Note that this fusion is only applicable when $\lambda^{fc}\neq0$ and $\lambda^{conv}\neq1$.

\section{Experimental Results}
We evaluate our approach
on MS COCO 2017 dataset \cite{lin2014microsoft} and Pascal VOC07 dataset \cite{everingham2010pascal}.
MS COCO 2017 dataset has 80 object categories. 
We train on \texttt{train2017} (118K images) and report results
on \texttt{val2017} (5K images) and  \texttt{test-dev} (41K images).
The standard COCO-style Average Precision (AP) with different IoU thresholds from 0.5 to 0.95 
is used as evaluation metric. 
Pascal VOC07 dataset has 20 object categories. We train on \texttt{trainval}
with 5K images and report results on \texttt{test} with 5K images. 
We perform ablation studies to analyze different components of our approach, and compare our approach to baselines and state-of-the-art.

\subsection{Implementation Details}
Our implementation is based on Mask R-CNN benchmark in Pytorch 1.0 \cite{massa2018mrcnn}.
Images are resized such that the shortest side is 800 pixels. We use no data augmentation for testing, and only horizontal flipping augmentation for training. The implementation details are described as follows:

\noindent \textbf{Architecture:} 
Our approach is evaluated on two FPN \cite{Lin_FPN} backbones (ResNet-50 and ResNet-101 \cite{he2016deep}), which are pretrained on {ImageNet} \cite{deng2009imagenet}. The standard RoI pooling is replaced by RoIAlign \cite{he2017mask}. Both heads and RPN are jointly trained end to end. All batch normalization (BN) \cite{ioffe2015batch} layers in the backbone are frozen. Each convolution layer in \textit{conv-head} is followed by a BN layer. The bounding box regression is class-specific.

\noindent \textbf{Hyper-parameters:} 
All models are trained using 4 NVIDIA P100 GPUs with 16GB memory, and a mini-batch size of 2 images per GPU. The weight decay is 1e-4 and momentum is 0.9. 

\noindent  \textbf{Learning Rate Scheduling:}
All models are  fine-tuned with 180k iterations. The learning rate is initialized with 0.01 and reduced by 10 after 120K and 160K iterations, respectively.

\subsection{Ablation Study}\label{sec:ablation:threecases}
We perform a number of ablations 
to analyze Double-Head with ResNet-50 backbone on COCO \texttt{val2017}. 

\noindent  \textbf{Double-Head Variations}: 
Four variations of double heads are compared:
\begin{itemize}
\setlength\itemsep{0em}
\item  \textbf{Double-FC} splits the classification and bounding box regression into two fully connected heads, which have the identical structure. 
\item \textbf{Double-Conv} splits the classification and bounding box regression into two convolution heads, which have the identical structure.
\item \textbf{Double-Head} includes a fully connected head (\textit{fc-head}) for classification and a convolution head (\textit{conv-head}) for bounding box regression.
\item \textbf{Double-Head-Reverse} switches tasks between two heads (i.e. \textit{fc-head} for bounding box regression and \textit{conv-head} for classification), compared to Double-Head. 
\end{itemize}{}

\begin{table}[t]
	\begin{center}
\begin{footnotesize}
		\begin{tabular}{c|c c c c|c c c|c c c}
\bottomrule
&\multicolumn{2}{|c}{\textit{fc-head}}   &	\multicolumn{2}{c|}{\textit{conv-head}} \\
			&cls & reg & cls & reg &AP  \\
			\hline
			%
			Single-FC  & 1.0 & 1.0 & - & - &36.8  \\
			Single-Conv & - & - & 1.0 & 1.0 & 35.9  \\
			\hline
			Double-FC   & 1.0 & 1.0 & - & - &{37.3}  \\
			Double-Conv & - & - & 1.0 & 1.0 & {33.8}  \\
			Double-Head-Reverse & - & 1.0 & 1.0 & - & {32.6} 	 \\
			Double-Head & 1.0 & - & - & 1.0 & \textbf{38.8} \\
			\hline
			Double-FC   & 2.0 & 2.0 & - & - &{38.1}  \\
			Double-Conv & - & - & 2.5 & 2.5 & {34.3}  \\
			Double-Head-Reverse & - & 2.0 & 2.5 & - & {32.0}	 \\
			Double-Head & 2.0 & - & - & 2.5 & \textbf{39.5}	 \\
			\toprule
		\end{tabular}
\end{footnotesize}
	\end{center}
	\caption{Evaluations of detectors with different head structures on COCO \texttt{val2017}. The backbone is FPN with ResNet-50. The top group shows performances for single head detectors. The middle group shows performances for detectors with double heads. The weight for each loss (classification and bounding box regression) is set to 1.0. Compared to the middle group, the bottom group uses different loss weight for \textit{fc-head} and \textit{conv-head} ($\omega^{fc}=2.0$, $\omega^{conv}=2.5$). 
	Clearly, Double-Head has the best performance, outperforming others by a non-negligible margin. Double-Head-Reverse has the worst performance. }
	\label{table:headcombine}
\end{table}

The detection performances are shown in Table \ref{table:headcombine}. The top group shows performances for single head detectors.  The middle group shows performances for detectors with double heads. The weight for each loss (classification and bounding box regression) is set to 1.0. Compared to the middle group, the bottom group uses different loss weights for \textit{fc-head} and \textit{conv-head} ($\omega^{fc}=2.0$ and $\omega^{conv}=2.5$), which are set empirically.

Double-Head outperforms single head detectors by a non-negligible margin (2.0+ AP). It also outperforms Double-FC and Double-Conv by at least 1.4 AP. Double-Head-Reverse has the worst performance (drops 6.2+ AP compared to Double-Head). This validates our findings that \textit{fc-head} is more suitable for classification, while \textit{conv-head} is more suitable for localization.

\begin{figure}[t]
	\begin{center}
        \includegraphics[width=\linewidth]{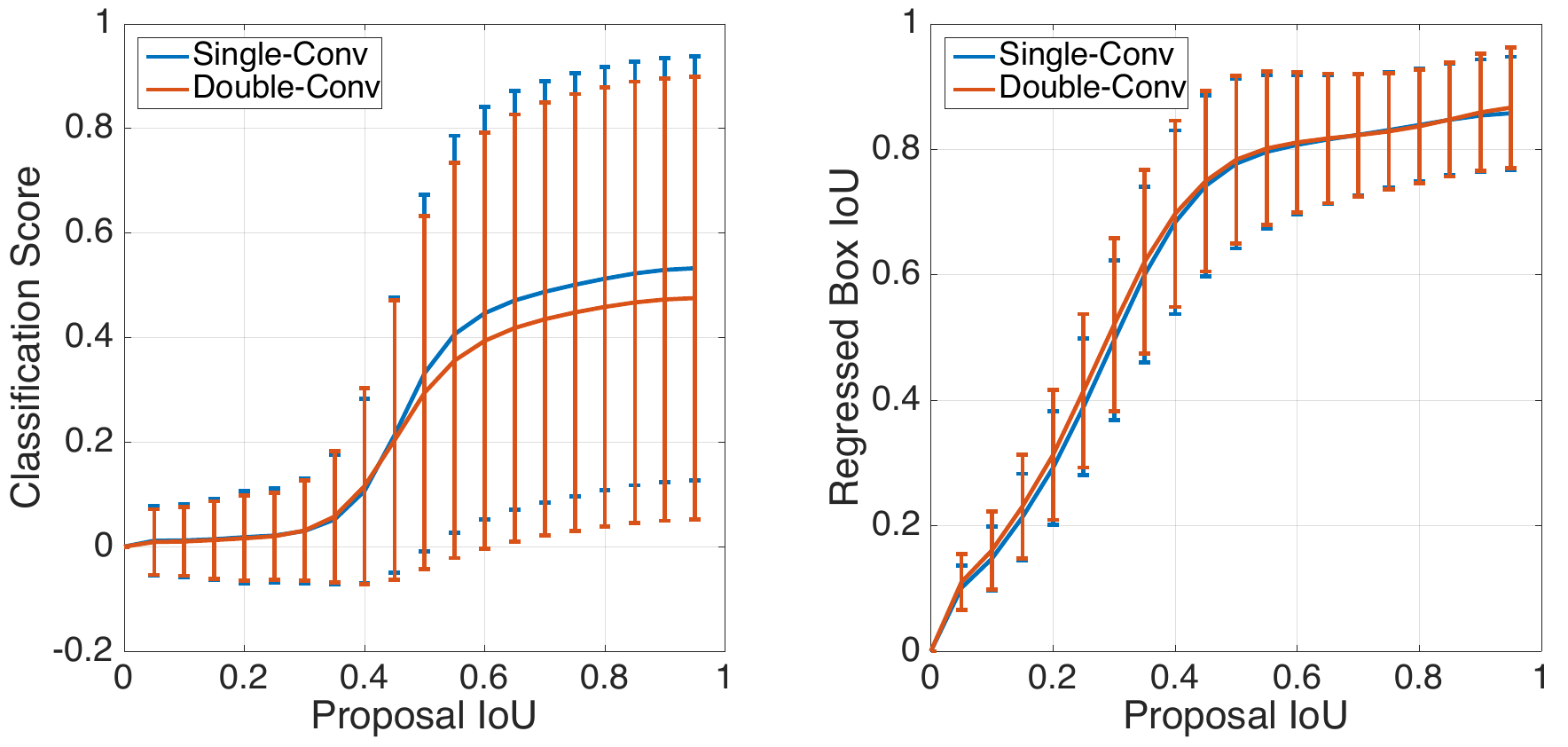}
	\end{center}
	\vspace{-3mm}
	\caption{
	 Comparison between  Single-Conv and Double-Conv. Left: mean and standard deviation of classification scores. Right: mean and standard deviation of IoUs between regressed boxes and their corresponding ground truth.
	Single-Conv has higher classification scores than Double-Conv, while regression results are comparable.
	}
	\label{fig:doubleConvComparision}
\end{figure}

Single-Conv performs better than Double-Conv. We believe that the regression task helps the classification task when sharing a single convolution head.
This is supported by sliding window analysis (see details in section \ref{sec:slidingwindowclassification}). Figure \ref{fig:doubleConvComparision} shows the comparison between Single-Conv and Double-Conv. Their regression results are comparable. 
But Single-Conv has higher classification scores than Double-Conv on proposals which have higher IoUs with the ground truth box.
Thus, sharing regression and classification on a single convolution head encourages the correlation between classification scores and proposal IoUs.
This allows Single-Conv to better determine if a complete object is covered.

\begin{figure}[!t]
	\begin{center}
        \includegraphics[width=0.95\linewidth]{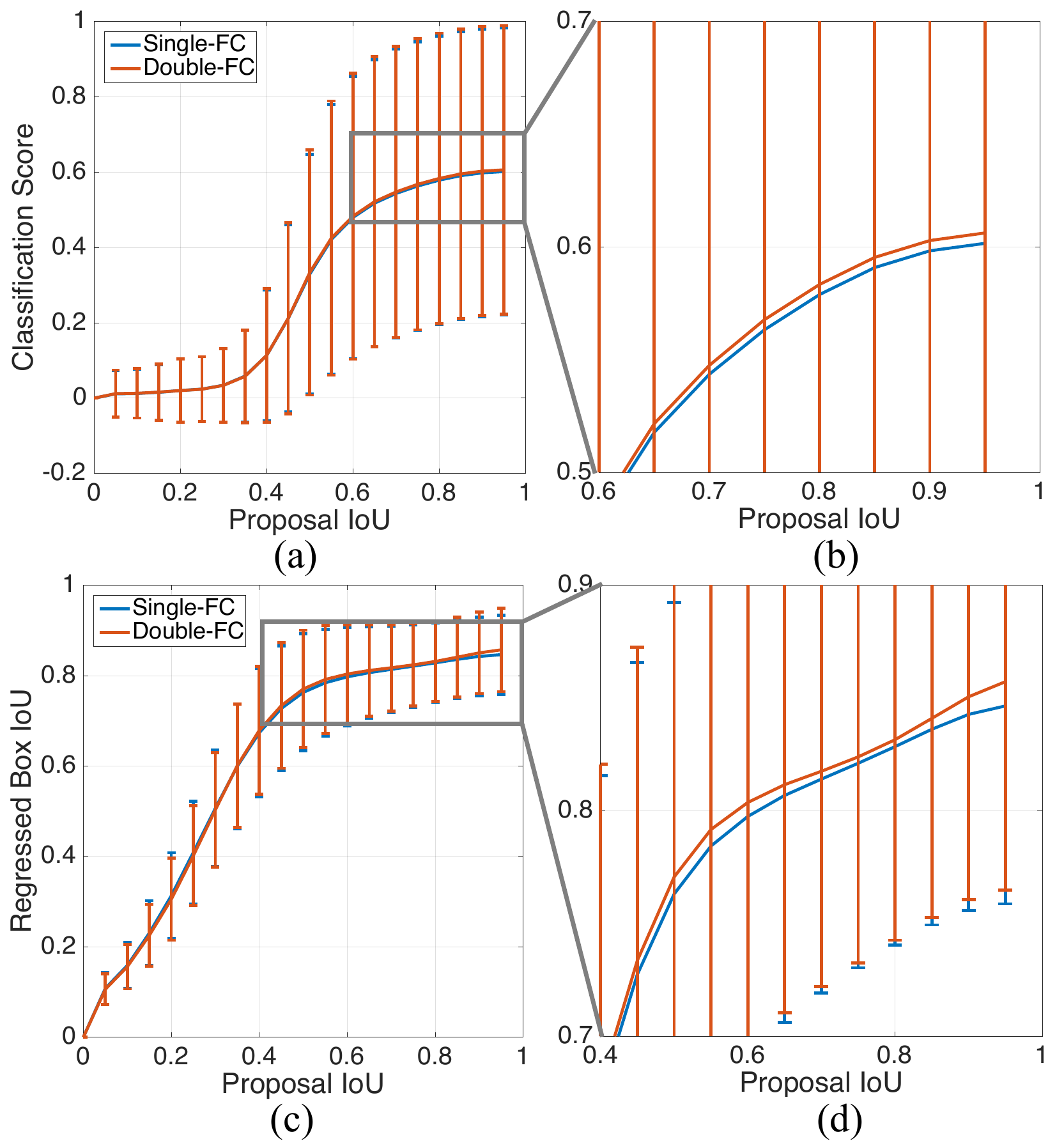}
	\end{center}
	\caption{
	Comparison between Single-FC and Double-FC. (a): mean and standard deviation of classification scores. (b): zooming in of the box in plot-(a). (c): mean and standard deviation of IoUs between regressed boxes and their corresponding ground truth. (d): zooming in of the box in plot-(c).
	Double-FC has slightly higher classification scores and better regression results than Single-FC.
 	}
	\label{fig:doubleFCComparision}
\end{figure}

In contrast, sharing two tasks in a single fully connected head (Single-FC) is not as good as separating them in two heads (Double-FC). 
We believe that adding the regression task in the same head with equal weight introduces confliction.
This is supported by the sliding window analysis.  Figure \ref{fig:doubleFCComparision} shows that Double-FC has slightly higher classification scores and higher IoUs between the regressed boxes and their corresponding ground truth than Single-FC.
\noindent  \textbf{Depth of \textit{conv-head}}: 
We study the number of blocks for the convolution head. The evaluations are shown in Table \ref{table:convolutionHeadStack}. The first group has $K$ residual blocks (Figure \ref{fig:upchannels}-(a-b)), while the second group has alternating $(K+1)/2$ residual blocks and $(K-1)/2$ non-local blocks (Figure \ref{fig:upchannels}-(c)). When using a single block in \textit{conv-head}, the performance is slightly behind FPN baseline (drops 0.1 AP) as it is too shallow. 
However, adding another convolution block boosts the performance substantially (gains 1.9 AP from FPN baseline). As the number of blocks increases, the performance improves gradually with decreasing growth rate. Considering the trade-off between accuracy and complexity, we choose \textit{conv-head} with 3 residual blocks and 2 non-local blocks ($K=5$ in the second group) for the rest of the paper, which gains 3.0 AP from baseline.


\noindent  \textbf{More training iterations}:
When increasing training iterations from 180k ($1 \times$ training) to 360k ($2 \times$ training), Double-Head gains 0.6 AP (from 39.8 to 40.4).

\begin{table}[!t]
	\begin{center}
\begin{footnotesize}
		\begin{tabular}{ c c|c|c c c }
            \bottomrule
			NL&$K$ & param &AP &	AP$_{0.5}$ & AP$_{0.75}$  \\
			\hline
			&0 & - & 36.8 &	58.7 & 40.4  \\
			\hline
			&1 & 1.06M & 36.7 { (-0.1)} & 	{59.3} &	{39.6} \\
			&2 & 2.13M & 38.7 (+1.9) & 	{59.2} &	41.9 \\
			&3 & 3.19M & {39.2 (+2.4)} & 	{59.4} &	{42.5}\\
			&4 & 4.25M & 39.3 (+2.5)&	{59.2} &	42.9 \\
			&5 & 5.31M & 39.5 (+2.7)&	{59.6} &	43.2 \\
			&6 & 6.38M & 39.5 (+2.7)&	{59.4} &	{43.3} \\
			&7 & 7.44M & 39.7 (+2.9)&	{59.8} &	43.2 \\
			\hline
			\checkmark&3 & 4.13M & 38.8 (+2.0) &	{59.2} &	42.4\\
			\checkmark&5 & 7.19M & 39.8 (+3.0) &	{59.6} &	43.6\\
			\checkmark&7 & 10.25M & \textbf{40.0 (+3.2)} &	\textbf{59.9} &	\textbf{43.7}\\
			\toprule
		\end{tabular}
\end{footnotesize}
	\end{center}
	\caption{The number of blocks (Figure \ref{fig:upchannels}) in the convolution head. The baseline ($K=0$) is equivalent to the original FPN \cite{Lin_FPN} which uses \textit{fc-head} alone. The first group only stacks residual blocks, while the second group alternates $(K+1)/2$ residual blocks and $(K-1)/2$ non-local blocks. 
	}
	\label{table:convolutionHeadStack}
\end{table}


\begin{table}[!t]
	\begin{center}
\begin{footnotesize}
		\begin{tabular}{ c | c c c c }
			\bottomrule
			Fusion Method  &AP &	AP$_{0.5}$ & AP$_{0.75}$ \\
			\hline
			No fusion & 39.7 & 	{59.5} &	43.4 \\
			Max & {39.9} & 	{59.7} &	{43.7}	\\
			Average & 40.1 & 	{59.8} &	44.1 \\
			Complementary & \textbf{40.3} & 	\textbf{60.3} &	\textbf{44.2}\\
			\toprule
		\end{tabular}
\end{footnotesize}
	\end{center}
	\caption{Fusion of classifiers from both heads. Complementary fusion (Eq. \ref{eq:cls-fusion}) outperforms others. The model is trained using weights $\lambda^{fc}=0.7$, $\lambda^{conv}=0.8$.}
	\label{table:cls-fusion}
\end{table}

\begin{figure*}[!t]
	\begin{center}
		\includegraphics[width=\linewidth]{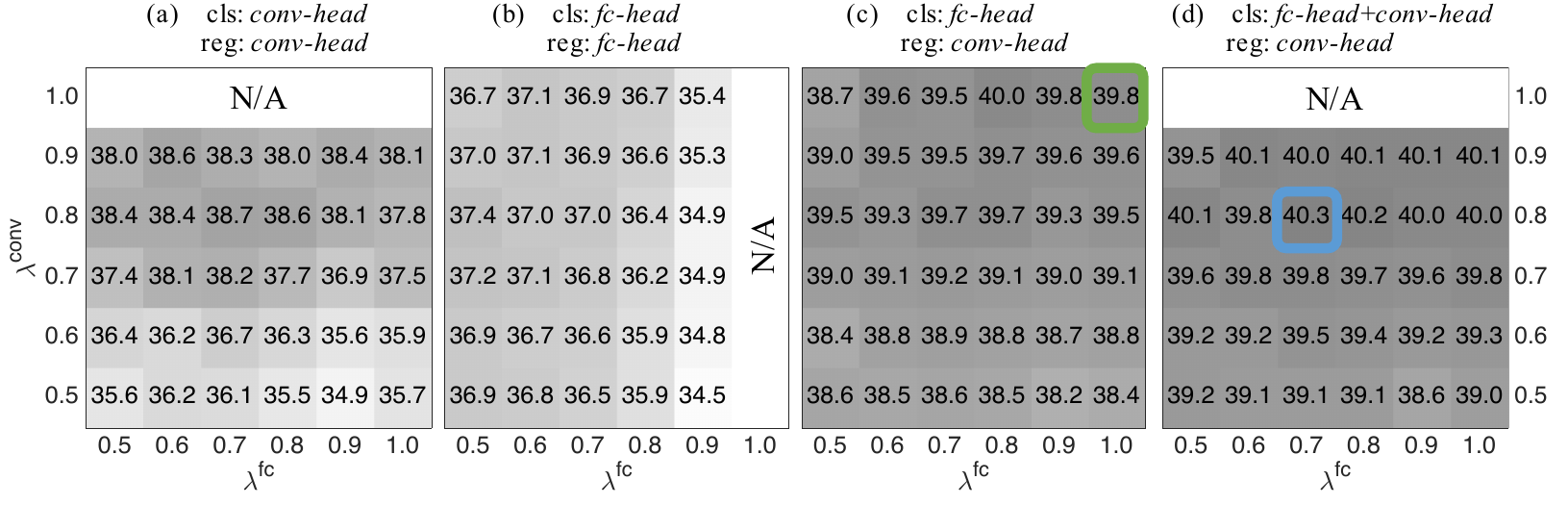}
	\end{center}
	\caption{AP over balance weights $\lambda^{fc}$ and $\lambda^{conv}$. For each ($\lambda^{fc}$, $\lambda^{conv}$) pair, we trained a Double-Head-Ext model. Note that the vanilla Double-Head is a special case with $\lambda^{fc}=1$, $\lambda^{conv}=1$.
		For each model, we evaluate AP in four ways: (a) using \textit{conv-head} alone, (b) using \textit{fc-head} alone, (c) using classification from \textit{fc-head} and bounding box from \textit{conv-head}, and (d) using classification fusion from both heads and bounding box from \textit{conv-head}. 
		Note that the first row in (a) and (d) is not available, due to the unavailability of classification in \textit{conv-head} when $\lambda^{conv}=1$. The last column in (b) is not available, due to the unavailability of bound box regression in \textit{fc-head} when $\lambda^{fc}=1$.}
	\label{fig:parameterSearch}
\end{figure*}

\begin{table} [!t]
	\begin{center}
\begin{footnotesize}
		\begin{tabular}{c |  c c c }
\bottomrule
			Method & AP & 	AP$_{0.5}$ & AP$_{0.75}$ \\ 
			\hline
			 FPN baseline \cite{Lin_FPN}  & {47.4} &	{75.7}&	{41.9} \\ 
			Double-Head-Ext (ours) & \textbf{49.2} &	\textbf{76.7}&	\textbf{45.6} \\ 
\toprule
		\end{tabular}
\end{footnotesize}
	\end{center}
	\caption{ Comparisons with FPN baseline \cite{Lin_FPN} on VOC07 datasets with ResNet-50 backbone. Our Double-Head-Ext outperforms FPN baseline. 
	}
	\label{table:comparisonBaselineVOC}
\end{table}

\begin{table*} [!t]
	\begin{center}
\begin{footnotesize}
		\begin{tabular}{c c |  c c c | c c c c c|c c c|c c c|c|c|c|c|c|c|}
			\bottomrule
			Method & Backbone  & AP & 	AP$_{0.5}$ & AP$_{0.75}$ & AP$_s$ & AP$_m$&	AP${_l}$ \\ 
			\hline
			Faster R-CNN \cite{ren2015faster} & ResNet-50-C4 & 34.8 & 	{55.8} &	37.0&		19.1&	38.8&	48.2  \\
			 FPN baseline \cite{Lin_FPN} & ResNet-50   &  36.8 &	58.7 & 40.4 & 21.2 & 40.1 & 48.8 \\
			 Double-Head (ours) & ResNet-50  & {39.8}&	{59.6}&	{43.6}&	\textbf{22.7}& {42.9}&	{53.1} \\
			Double-Head-Ext (ours) & ResNet-50 &  \textbf{40.3}&	\textbf{60.3}&	\textbf{44.2}& {22.4}&	\textbf{43.3}&	\textbf{54.3} \\ 
			\hline
			 Faster R-CNN \cite{ren2015faster} & ResNet-101-C4 &  38.5 & 	{59.4} &	41.4&		19.7&	43.1&	53.3  \\
			 FPN baseline \cite{Lin_FPN}  & ResNet-101 & {39.1}	&61.0&	{42.4} &		{22.2}&	42.5&	51.0  \\
			 Double-Head (ours) & ResNet-101 & {41.5} &	{61.7} &	{45.6} & {23.8} &	\textbf{45.2}&	{54.9} \\
			Double-Head-Ext (ours) & ResNet-101&  \textbf{41.9} &	\textbf{62.4} &	\textbf{45.9} &		\textbf{23.9} &	\textbf{45.2}&	\textbf{55.8} \\
\toprule
		\end{tabular}
\end{footnotesize}
	\end{center}
	\caption{Object detection results (bounding box AP) 
	on COCO \texttt{val2017}. Note that FPN baseline only has \textit{fc-head}. Our Double-Head and Double-Head-Ext outperform both Faster R-CNN and FPN baselines on two backbones (ResNet-50 and ResNet-101). 
	}
	\label{table:comparisonBaseline}
\end{table*}

\noindent  \textbf{Balance Weights $\lambda^{fc}$ and $\lambda^{conv}$}: Figure \ref{fig:parameterSearch} shows APs for different choices of $\lambda^{fc}$ and $\lambda^{conv}$. For each ($\lambda^{fc}$, $\lambda^{conv}$) pair, we train a Double-Head-Ext model. The vanilla Double-Head model is corresponding to $\lambda^{fc}=1$ and $\lambda^{conv}=1$, while other models involve supervision from unfocused tasks. For each model, we evaluate AP for using \textit{conv-head} alone (Figure \ref{fig:parameterSearch}-(a)), using \textit{fc-head} alone (Figure \ref{fig:parameterSearch}-(b)), using classification from \textit{fc-head} and bounding box from \textit{conv-head} (Figure \ref{fig:parameterSearch}-(c)), and using classification fusion from both heads and bounding box from \textit{conv-head} (Figure \ref{fig:parameterSearch}-(d)). $\omega^{fc}$ and $\omega^{conv}$ are set as $2.0$ and $2.5$ in all experiments, respectively.

We summarize key observations as follows. Firstly, using two heads (Figure \ref{fig:parameterSearch}-(c)) outperforms using a single head (Figure \ref{fig:parameterSearch}-(a), (b)) for all ($\lambda^{fc}$, $\lambda^{conv}$) pairs by at least 0.9 AP. 
Secondly, fusion of classifiers introduces at least additional 0.4 AP improvement for all ($\lambda^{fc}$, $\lambda^{conv}$) pairs (compare Figure \ref{fig:parameterSearch}-(c) and (d)).
%
%
And finally the unfocused tasks are helpful as the best Double-Head-Ext model (40.3 AP) is corresponding to $\lambda^{fc}=0.7$, $\lambda^{conv}=0.8$ (blue box in Figure \ref{fig:parameterSearch}-(d)). It outperforms Double-Head (39.8 AP, green box in Figure \ref{fig:parameterSearch}-(c)) without using unfocused tasks by 0.5 AP. For the rest of the paper, we use $\lambda^{fc}=0.7$ and $\lambda^{conv}=0.8$ for Double-Head-Ext.


\noindent  \textbf{Fusion of Classifiers}: We study three different ways to fuse the classification scores from both the fully connected head ($s^{fc}$) and the convolution head ($s^{conv}$) during inference: (a) average, (b) maximum, and (c) complementary fusion using Eq. (\ref{eq:cls-fusion}). 
The evaluations are shown in Table \ref{table:cls-fusion}. The proposed complementary fusion outperforms other fusion methods (max and average) and gains 0.6 AP compared to using the score from \textit{fc-head} alone. 

\begin{table*} [!t]
	\begin{center}
\begin{footnotesize}
		\begin{tabular}{c c |  c c c | c c c c c|c c c|c c c|c|c|c|c|c|c|}
			\bottomrule
			Method & Backbone  & AP & 	AP$_{0.5}$ & AP$_{0.75}$ & AP$_s$ & AP$_m$&	AP${_l}$ \\ 
			\hline
            FPN \cite{Lin_FPN} & ResNet-101  & 36.2 & 59.1 &  39.0 & 18.2 & 39.0 & 48.2 \\ 
            Mask RCNN  \cite{he2017mask} & ResNet-101  & 38.2 & 60.3 & 41.7 & 20.1 & 41.1 & 50.2 \\
            Deep Regionlets \cite{Xu_2018_ECCV}&  ResNet-101 & 39.3 &  59.8 &  - & 21.7 &  43.7 & 50.9\\
            IOU-Net \cite{jiang2018acquisition} & ResNet-101 & 40.6 & 59.0 &  - & - & - &- \\ 
            Soft-NMS \cite{bodla2017soft} & Aligned-Inception-ResNet & 40.9& \textbf{62.8} & - & 23.3 & 43.6 & 53.3\\
			LTR \cite{Tan_2019_ICCV} & ResNet-101 & {41.0}&	{60.8}&	{44.5}&		{23.2}&	{44.5}&	{52.5} \\ 
            Fitness NMS \cite{tychsen2018improving} & DeNet-101 \cite{tychsen2017denet} &  41.8 &  60.9 & 44.9 & 21.5 &\textbf{45.0}  & \textbf{57.5} \\
			Double-Head-Ext (ours) & ResNet-101 & \textbf{42.3}&	\textbf{62.8}&	\textbf{46.3}&		\textbf{23.9}&	{44.9}&	{54.3} \\ 
			\toprule
		\end{tabular}
\end{footnotesize}
	\end{center}
	\caption{ 
	Object detection results (bounding box AP), \textit{vs.} state-of-the-art on COCO \texttt{test-dev}. 
	All methods are in the family of two-stage detectors with a single training stage. 
	Our Double-Head-Ext achieves the best performance.
	}
	\label{table:comparisonSOTA}
\end{table*}

\subsection{Main Results}

\noindent  \textbf{Comparison with Baselines on VOC07}:
We conduct experiments on Pascal VOC07 dataset and results are shown in Table \ref{table:comparisonBaselineVOC}. Compared with FPN, our method gains 1.8 AP. Specifically, it gains 3.7 AP$_{0.75}$ on the higher IoU threshold 0.75 and gains 1.0 AP$_{0.5}$ on the lower IoU threshold 0.5. 
\noindent  \textbf{Comparison with Baselines on COCO}:
Table \ref{table:comparisonBaseline} shows the comparison between our method 
with Faster RCNN \cite{ren2015faster} and FPN \cite{Lin_FPN} baselines on COCO \texttt{val2017}.
%
Our method outperforms both baselines on \textit{all} evaluation metrics. Compared with FPN, our Double-Head-Ext gains 3.5 and 2.8 AP on ResNet-50 and ResNet-101 backbones, respectively. Specifically, our method gains 3.5+ AP on the higher IoU threshold (0.75) and 1.4+ AP on the lower IoU threshold (0.5) for both backbones. This demonstrates the advantage of our method with double heads. 

We also observe that Faster R-CNN and FPN have different preferences over object sizes when using ResNet-101 backbone: i.e. Faster R-CNN has better AP on medium and large objects, while FPN is better on small objects. Even comparing with the best performance among FPN and Faster R-CNN across different sizes, our Double-Head-Ext gains 1.7 AP on small objects, 2.1 AP on medium objects and 2.5 AP on large objects. This demonstrates the superiority of our method, which leverages the advantage of \textit{fc-head} on classification and the advantage of \textit{conv-head} on localization. 
\noindent  \textbf{Comparison with State-of-the-art on COCO}:
We compare our Double-Head-Ext with the state-of-the-art methods on MS COCO 2017 \texttt{test-dev} in Table \ref{table:comparisonSOTA}. ResNet-101 is used as the backbone. 
For fair comparison, the performance of single-model inference is reported for all methods.
Here, we only consider the two-stage detectors with a single training stage. 
Our Double-Head-Ext achieves the best performance with 42.3 AP. This demonstrates the superior performance of our method.
Note that Cascade RCNN \cite{Cai_2018_CVPR} is not included as it involves multiple training stages. 
Even through our method only has one training stage, the performance of our method is slightly below Cascade RCNN (42.8 AP).

\section{Conclusions}


In this paper, we perform a thorough analysis and find an interesting fact that two widely used head structures (convolution head and fully connected head) have {opposite} preferences towards classification and localization tasks in object detection. Specifically,  \textit{fc-head} is more suitable for the classification task, while \textit{conv-head} is more suitable for the localization task. 
Furthermore, we examine the output feature maps of both heads and find that \textit{fc-head} has more spatial sensitivity than \textit{conv-head}.
Thus, \textit{fc-head} has more capability to distinguish a complete object from part of an object, but is not robust to regress the whole object.
Based upon these findings, we propose a {Double-Head} method, which has a fully connected head focusing on classification and a convolution head for bounding box regression.  Without bells and whistles, our method gains +3.5 and +2.8 AP on MS COCO dataset from FPN baselines with ResNet-50 and ResNet-101 backbones, respectively.
We hope that our findings are helpful for future research in object detection.

\clearpage
\appendix
\section{APPENDIX}

In this appendix, we firstly compare \textit{fc-head}  and \textit{conv-head} on easy, medium and hard classes using sliding window analysis. Then, we examine the effect of joint training for both heads in Double-Head-Ext. Finally, we provide qualitative analysis to compare \textit{fc-head}, \textit{conv-head} and Double-Head-Ext.
\subsection{Sliding Window Analysis on Different Difficulty Levels of Classes}
In addition to the sliding window analysis on different sizes of objects
in the main paper, we perform the analysis based upon another criteria the difficulty of the object category. We rank all object classes based upon AP results of the FPN \cite{Lin_FPN} baseline. Each group on difficulty of the object category (easy/medium/hard) takes one third of classes. The results are shown in Figure \ref{fig:APPENDIXslidingwindowMeanstdDifficulty}.
Similarly, classification scores of \textit{fc-head} are more correlated to IoUs between proposals and corresponding ground truth than of \textit{conv-head}, especially for hard objects.  
The corresponding Pearson correlation coefficients (PCC) are shown in Figure \ref{fig:APPENDIXslidingwindowPscoresDifficulty}.
And the regressed boxes of \textit{conv-head} are more accurate when the proposal IoU is above 0.4.

\subsection{Effect of Joint Training in Double-Head-Ext}
To examine the effect of joint training with both heads, we conduct the sliding window analysis (details in section \ref{sec:slidingwindowclassification}) on Double-Head-Ext with the ResNet-50 backbone and compare the results on all objects with two single head models. Results in Figure \ref{fig:APPENDIXcomparisonSingleDouble} show: (i) the classification scores from \textit{conv-head} in Double-Head-Ext are improved significantly compared with using \textit{conv-head} alone and (ii) regression results from \textit{conv-head} of Double-Head-Ext are better as well.
We believe that these improvements come from the jointly training of two heads with the shared backbone.

\setcounter{figure}{0}
\renewcommand{\thefigure}{A.\arabic{figure}}

\begin{figure}[t]
	\begin{center}
        \includegraphics[width=\linewidth]{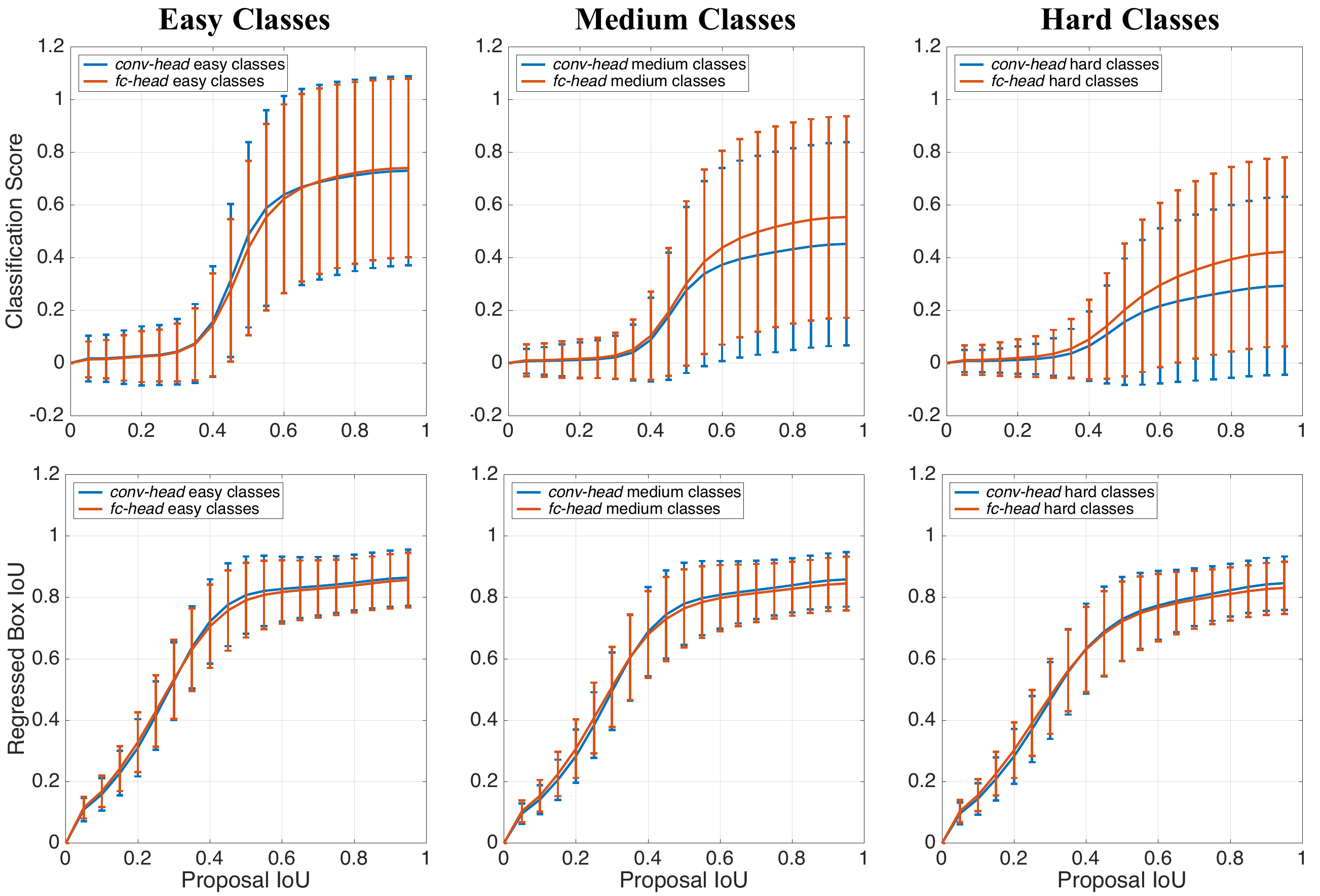}
	\end{center}
	\vspace{-2mm}
	\caption{
	Comparison between \textit{fc-head} and \textit{conv-head} on easy, medium and hard classes. Top row: mean and standard deviation of classification scores. Bottom row: mean and standard deviation of IoUs between regressed boxes and their corresponding ground truth. 
}
	\vspace{-2mm}
	\label{fig:APPENDIXslidingwindowMeanstdDifficulty}
\end{figure}

\begin{figure}[t]
	\begin{center}
		\includegraphics[width=0.45\linewidth]{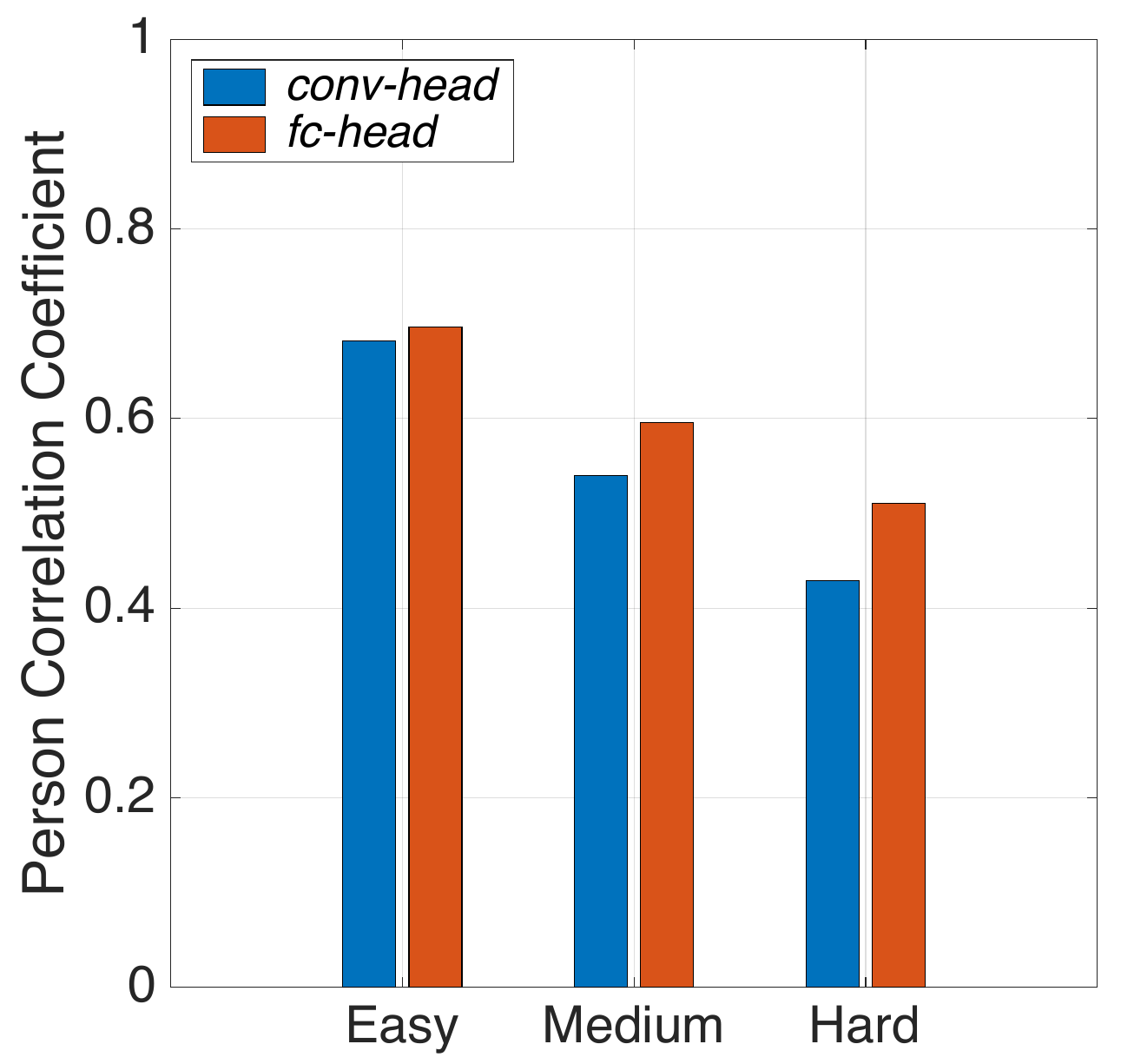}
	\end{center}
	\vspace{-2mm}
	\caption{
Pearson correlation coefficient (PCC) between classification scores and IoUs of predefined proposals for easy, medium and hard classes.
	}
	\vspace{-2mm}
	\label{fig:APPENDIXslidingwindowPscoresDifficulty}
\end{figure}

\begin{figure}[t]
	\begin{center}
		\includegraphics[width=0.8\linewidth]{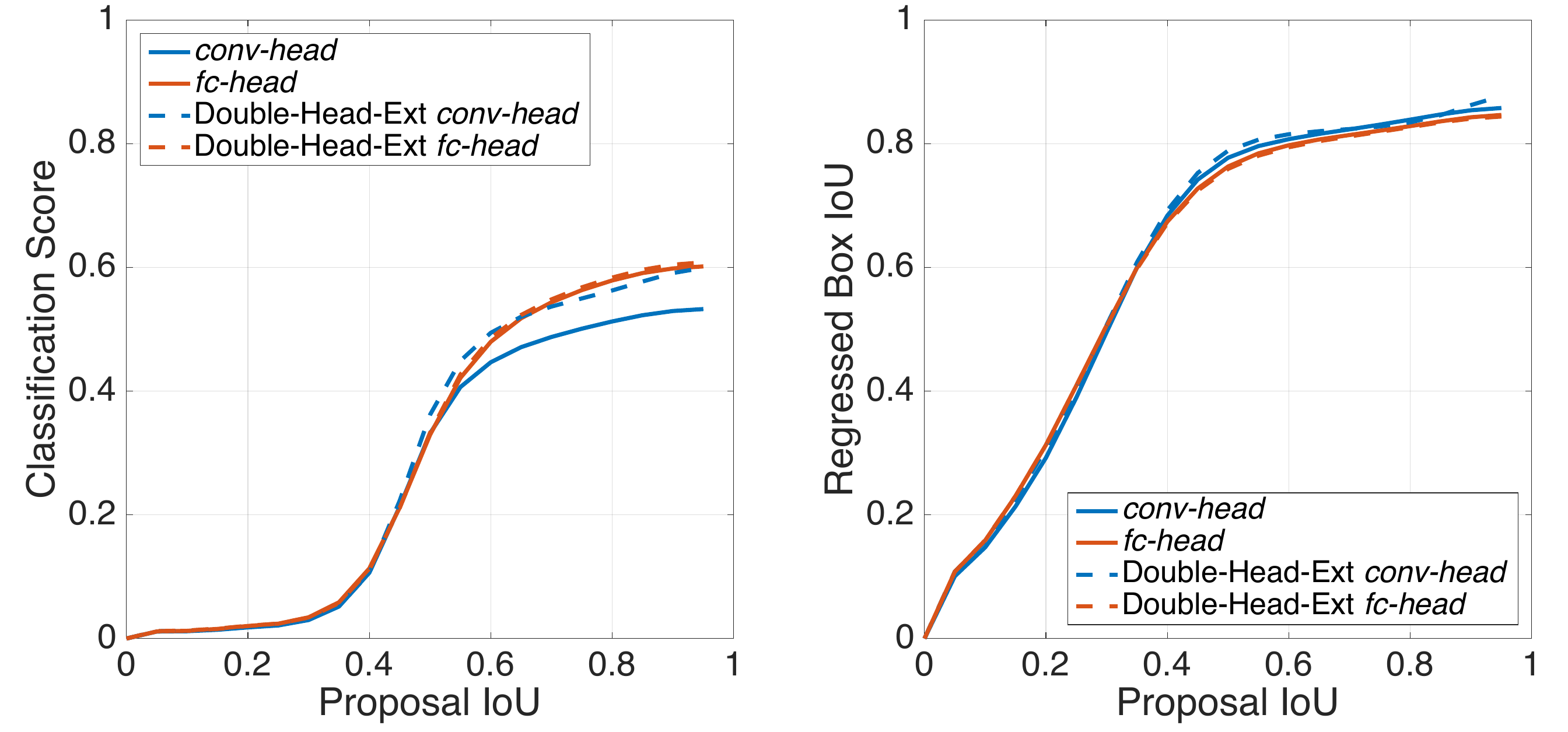}
	\end{center}
	\vspace{-2mm}
	\caption{Comparison of classification and bounding box regression. Left: mean of classification scores between single heads and Double-Head-Ext. Right: mean of regressed box IoUs between single heads and Double-Head-Ext.
	}
	\vspace{-4mm}
	\label{fig:APPENDIXcomparisonSingleDouble}
\end{figure}

\subsection{Qualitative Analysis}
We apply a well trained Double-Head-Ext model (Figure 1-(d) in the submission draft) 
and compare the detection results of (a) using \textit{conv-head} alone, (b) using \textit{fc-head} alone, and (c) using both heads.

\begin{figure*}[b]
	\begin{center}
		\includegraphics[width=0.8\linewidth]{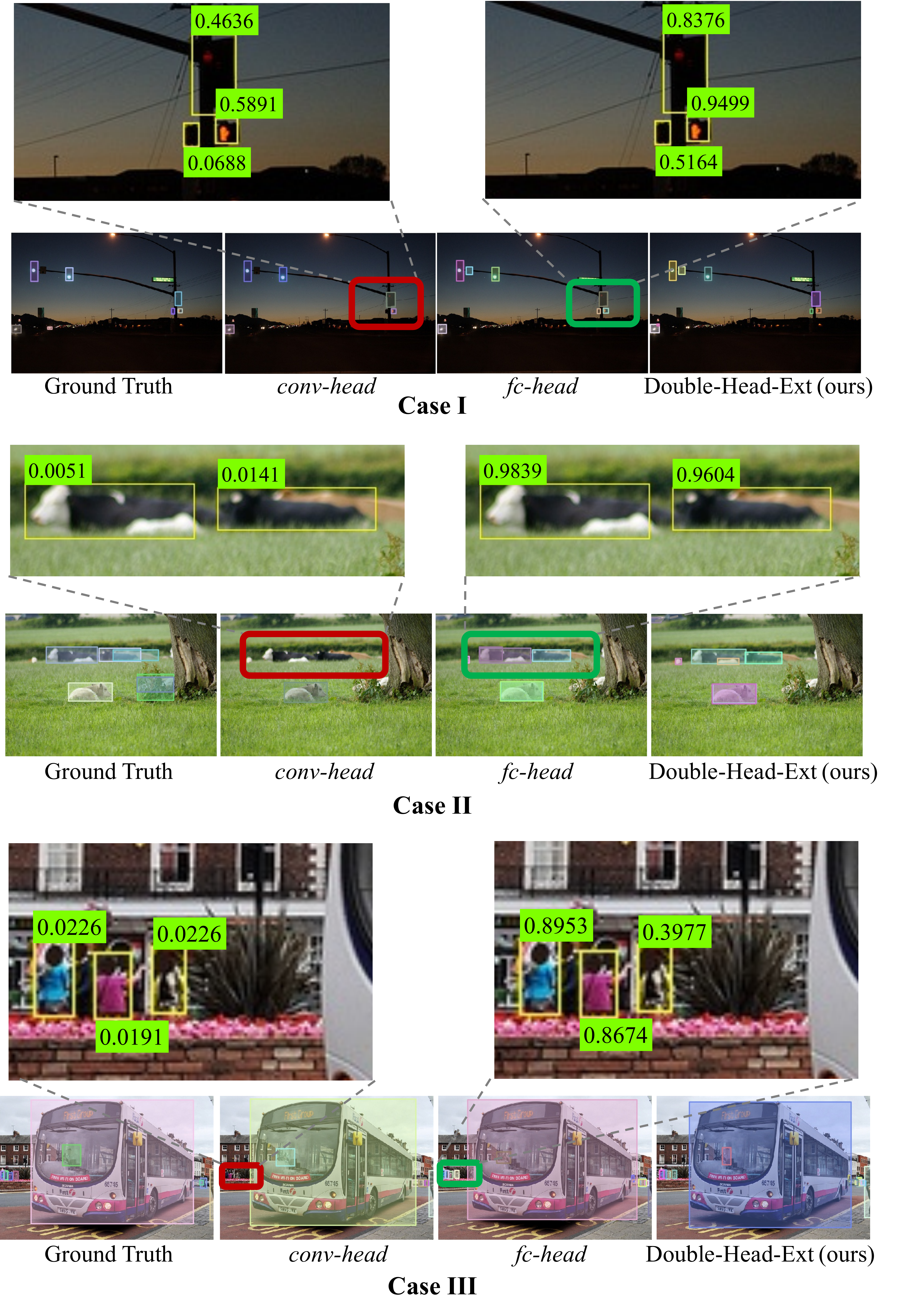}
	\end{center}
	\caption{\textit{fc-head} head is more suitable for classification than \textit{conv-head}. This figure includes three cases. Each case has two rows. The bottom row shows ground truth, detection results using \textit{conv-head} alone, \textit{fc-head} alone, and our Double-Head-Ext (from left to right). The \textit{conv-head} misses objects in the \textcolor{red}{red} box. In contrast, these missing objects are successfully detected by \textit{fc-head} (shown in the corresponding \textcolor{green}{green} box). The top row zooms in the red and green boxes, and shows classification scores from the two heads for each object. The missed objects in \textit{conv-head} have small classification scores, compared to \textit{fc-head}.}
	\label{fig:missV4extra}
\end{figure*}

Figure \ref{fig:missV4extra} shows three cases that \textit{fc-head} is better than \textit{conv-head} in classification. In all three cases, \textit{conv-head} misses small objects due to low classification scores (e.g. signal light in case I, cows in case II, and persons in case III). In contrast, these objects are successfully detected by \textit{fc-head} (in the green box) with proper classification scores. Our Double-Head-Ext successfully detects these objects, by leveraging the superiority of \textit{fc-head} for classification.

\begin{figure*}[!t]
	\begin{center}
		\includegraphics[width=\linewidth]{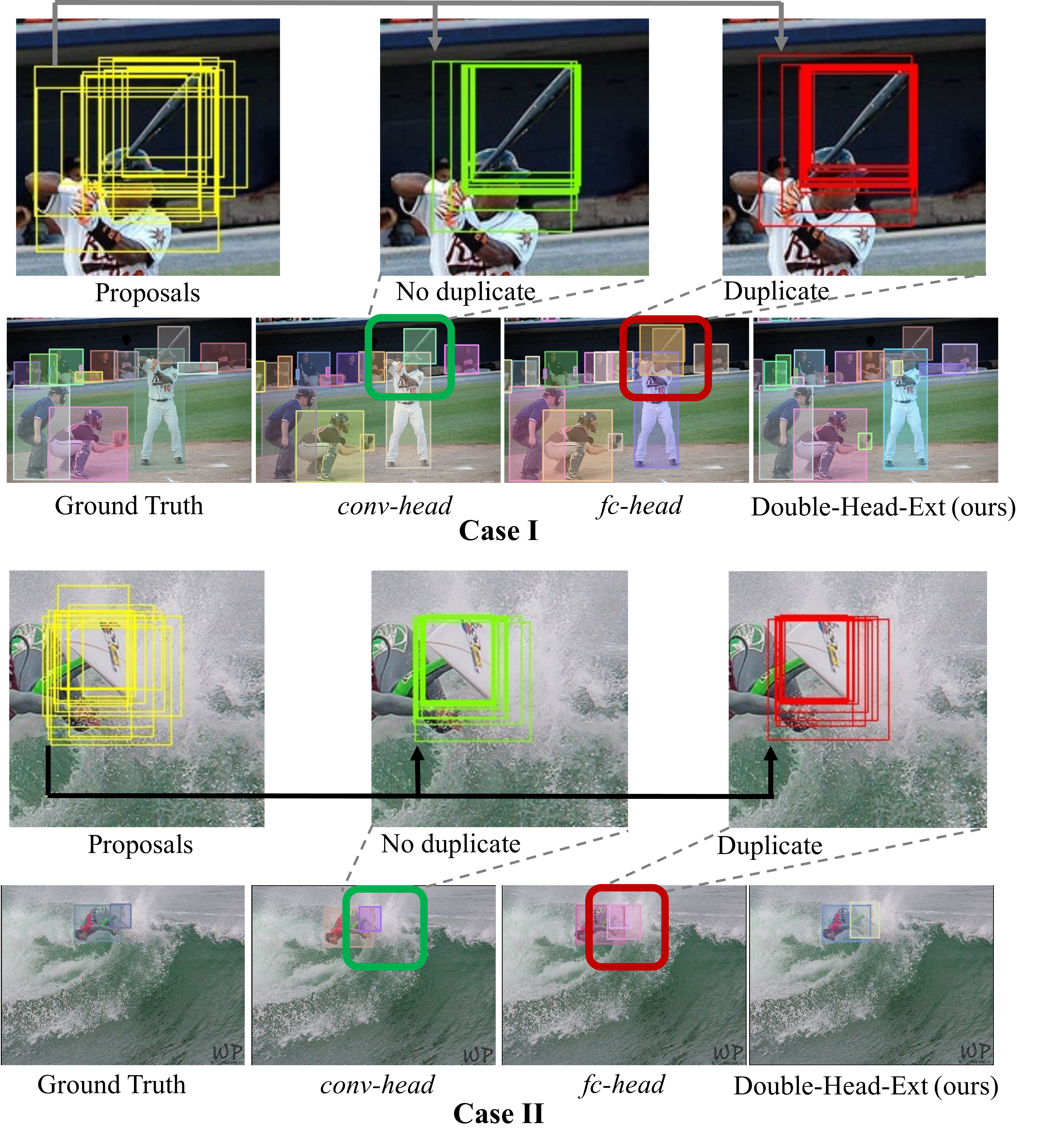}
	\end{center}
	\caption{
	\textit{conv-head} is more suitable for localization than \textit{fc-head}. This figure includes two cases. Each case has two rows. The bottom row shows ground truth, detection results using \textit{conv-head} alone, \textit{fc-head} alone, and our Double-Head-Ext (from left to right). \textit{fc-head} has a duplicate detection for the baseball bat in case I and for the surfing board in case II (in the red box). The duplicate detection is generated from an inaccurate proposal (shown in the top row), but is not removed by NMS due to its low IoU with other detection boxes. In contrast, \textit{conv-head} has more accurate box regression for the same proposal, with higher IoU with other detection boxes. Thus it is removed by NMS, resulting no duplication. }
	\label{fig:surfboard}
\end{figure*}

Figure \ref{fig:surfboard} demonstrates two cases that \textit{conv-head} is better than \textit{fc-head} in localization. 
Compared with \textit{fc-head} which has a duplicate detection for both cases (i.e. baseball bat in case I, and surfing board in case II, shown in the red box at the bottom row), \textit{conv-head} has a single accurate detection (in the green box at the bottom row). Both heads share proposals (yellow boxes). The duplicated box from \textit{fc-head} comes from an inaccurate proposal. It is not suppressed by NMS as it has low IoU with other boxes around the object. In contrast, \textit{conv-head} has a more accurate regression box for the same proposal, which has higher IoU with other boxes. This allows NMS to remove it. Double-Head-Ext has no duplication, by leveraging the superiority of \textit{conv-head} in localization.

\clearpage

\balance
{\small
\bibliographystyle{ieee_fullname}
\bibliography{egbib}
}

\end{document}